# LEFT VENTRICLE SEGMENTATION BY MODELLING UNCERTAINTY IN PREDICTION OF DEEP CONVOLUTIONAL NEURAL NETWORKS AND ADAPTIVE THRESHOLDING INFERENCE

*Alireza Norouzi[1], Ali Emami[2], S.M.Reza Soroushmehr[3, 4],*
*Nader Karimi[1], Shadrokh Samavi[1, 3], Kayvan Najarian[3, 4]*

[1]Department of Electrical and Computer Engineering, Isfahan University of Technology, Isfahan 84156-83111, Iran
[2]Department of Information Technology and Electrical Engineering, University of Queensland, Australia
[3]Department of Computational Medicine and Bioinformatics, University of Michigan, Ann Arbor, MI, U.S.A.
[4]Michigan Center for Integrative Research in Critical Care, University of Michigan, Ann Arbor, MI, U.S.A.

**ABSTRACT**

Deep neural networks have shown great achievements in solving complex problems. However, there are fundamental problems that limit their real world applications. Lack of measurable criteria for estimating uncertainty in the network outputs is one of these problems. In this paper, we address this limitation by introducing deformation to the network input and measuring the level of stability in the network's output. We calculate simple random transformations to estimate the prediction uncertainty of deep convolutional neural networks. For a real use-case, we apply this method to left ventricle segmentation in MRI cardiac images. We also propose an adaptive thresholding method to consider the deep neural network uncertainty. Experimental results demonstrate state-of-the-art performance and highlight the capabilities of simple methods in conjunction with deep neural networks.

## 1. INTRODUCTION

Cardiac diseases are one of the main causes of death especially in developed nations around the world. Despite massive investments on developing equipment, medicines and pre-caution strategies, there is still a huge gap between annual death reports caused by cardiac failures and an ideal world in which these diseases are fully under control [1]. In this situation, early detection of such life-threatening diseases is vital. High technology imaging systems, such as magnetic resonance imaging (MRI), boosted with advanced image processing and pattern recognition algorithms, have greatly improved the health care systems by introducing automatic medical diagnosis. Although human supervision is still needed due to high sensitivity of the field, the medical diagnosis process has been accelerated thanks to relatively high accuracy of the introduced automatic recognition systems.

The explosion of interest in using deep neural networks, fueled by their success in solving complex real world problems, has led to the birth of impressive automated medical image analysis systems. Despite their capacity to learn rich hierarchical features, their output is usually a single number, resembling a kind of belief the network has about the input. However, there is usually no clue about how confident the model is about its prediction. This could be an alarm, especially in critical applications such as medical diagnosis. In other words, we need to have an estimate about reliability or confidence of the output believes. In this work, we try to model uncertainty of the output probability map using Monte Carlo sampling from a linear manifold on which input images lie. This process is performed by applying a random affine transformation on input images. In contrast to other approaches which sample outputs by direct injection of noise to model internal representation [2], we perturb the input and check if the system can respond properly or not. In other words, we evaluate the system strength in performing a proper segmentation on deformed inputs. Since we know the perturbation operation and its effect on the input, our approach offers more interpretability and control over the proposed method for uncertainty evaluation. Having computed the heat maps and uncertainty of the model using this technique, we apply an adaptive thresholding method to get the final segmentation mask.

To summarize, the contribution of this paper is threefold: First, we compute uncertainty of the model output using Monte Carlo sampling in the input space using well-defined affine transformation. Second, we improve the model prediction and training process by defining a new loss function to improve gradient flow. Third, we propose an adaptive thresholding scheme to compute the final result.

The rest of the paper is organized as follows. In section 2, a literature review is presented and section 3 gives a detailed explanation of the proposed methods. In section 4, we discuss experimental results and finally we conclude the paper in section 5.

## 2. LITERATURE REVIEW

Invention of fully convolutional neural networks (FCNs) [3] paved the way for popularizing the use of deep neural networks in semantic segmentation tasks. Since then, many enhancements and architectural changes have been proposed for improving accuracy [4] and operational speed [5].

In the field of computational medical image analysis, a specific architecture called U-Net [6] is successfully applied for solving a range of complex tasks, including segmentation. Recently, a number of research papers have demonstrated the capability of deep learning methods in solving medical tasks comparable to that of human performance [7], [8].

Particularly, for the problem of left ventricle segmentation, researchers have proposed a wide range of methods. Algorithms based on active contours and shape models are arguably one of the earliest and most popular ones employed in this context [1]. More recently, a method based on dynamic programming has been proposed to segment the left ventricle in cardiac MRI [9]. Deep neural networks is one of the recent proposals in solving left ventricle segmentation problem [10]. Combining the ideas from deep convolutional neural networks and image processing techniques such as utilizing Gabor filters to initialize the filter weights in the network, has shown to be promising [11]. In this work, we try to achieve a good balance between the model accuracy and computational burdens, both in training and test phases.

For incorporating model uncertainty, one suggestion is to inject noise into network internal representations by various means such as test time dropout [2]. Another line of work is to make the network model a distribution family over the input rather than direct prediction of the desired output [12]. Compared to previous methods, our approach is certainly more controllable and interpretable, due to our knowledge of input space and perturbation process. However, these methods are complementary and one can easily combine them.

Adaptive thresholding is a mature topic in classical image processing. Otsu and Sauvola [13] are among the most popular methods to name a few. We have borrowed some intuitions from these works on our adaptive thresholding formulization.

## 3. PROPOSED METHOD

The proposed pipeline consists of three major components working in an iterative manner. First of all, we use a random transformation generator which takes an image as input and generates several random transformations. The second component at the core of the pipeline is a deep convolutional neural network based on U-Net [6] and ResNet [14] architectures. This module generates a heat map for the transformed input. These two components of the pipeline work in a loop for a specified number of iterations in order to generate a batch of heat maps for any input image. The last module in the pipeline uses these outputs to compute appropriate statistics for the adaptive threshold computation, which is used for predicting the final segmentation map. The whole pipeline is shown in figure

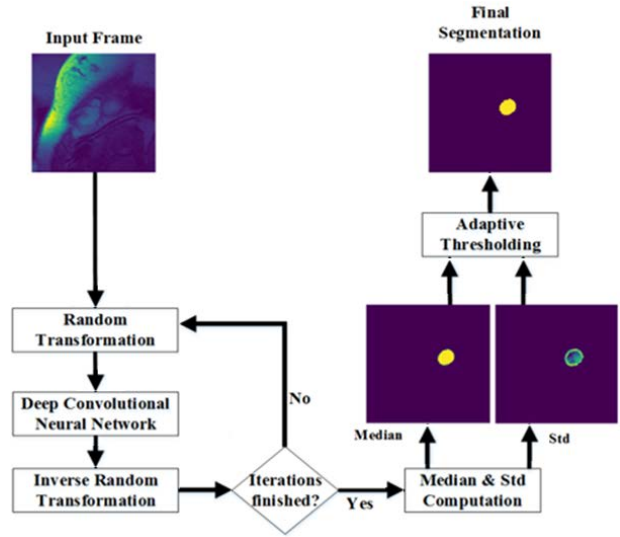

Figure 1. Proposed pipeline.

(1). In the following subsections, we elaborate on details of each part.

### 3.1. Random transformation generator

There is a variety of ways to transform an image. It can be based on geometric operations such as scaling, translation, rotation and affine transformation. There are also methods based on the intensity values, e.g. contrast jitter or intensity histogram mappings. In this work, we limit out approach to geometric operations. For this purpose, a random affine transformation is applied on each input image before feeding to the deep convolutional neural network. The random affine matrix $A$, is built based on the following distributions:

$$A(t_x, t_y, \theta) = \begin{bmatrix} \cos(\theta) & -\sin(\theta) & t_x \\ \sin(\theta) & \cos(\theta) & t_y \\ 0 & 0 & 1 \end{bmatrix} \quad (1)$$

$$t_x, t_y \sim Uniform(-\frac{T}{2}, +\frac{T}{2}) \quad (2)$$

$$\theta \sim Uniform(-\frac{\pi R}{360}, +\frac{\pi R}{360}) \quad (3)$$

where scalars $T$ and $R$ determine the range of translation (in pixels) and rotation (in degrees) respectively. As is evident from (1), we only use translation and rotation transformations to build the affine matrix. Each generated matrix is stored in memory to compute the inverse transformation in later stages. Another possible extension to this line of work is to consider a scaling factor in matrix $A$ or equivalently apply matrix $A$ on a scale pyramid of the input image. However, in this work, we are focused on modeling uncertainty of deep CNNs by predicting the output in presence of simple affine transformations. More complex transformations may be investigated in future works.

## 3.2. Heat-map prediction

The core of the pipeline is built upon fully convolutional neural network (CNN) idea. Among vast variety of architectures, U-Net is one of the most well-known ones for medical image analysis. Due to the great success of deep learning in general and specifically U-Net architecture in solving complex medical tasks [6], we build our CNN based on the idea of U-Net and improve it using the most recent techniques from deep learning literature. Briefly, U-Net is an auto-encoder based architecture in which the encoder extracts the most salient features concerning the input-output relationship. Given the encoded features of input images, decoder tries to predict the final answer, which in our case is a segmentation map. Furthermore, there are data flow connections between each encoder and its corresponding decoder. These shortcut links are crucial, especially in problems such as semantic segmentation to preserve spatial information which might be corrupted as a result of down-sampling procedures (e.g. max-pooling or strided convolution) [15]. As mentioned earlier, we borrow the main ideas of U-Net architecture but decompose it into smaller sub-modules in order to design the new model more systematically. For this purpose, we split encoder and decoder into smaller building blocks. Each sub-module consists of smaller blocks containing consecutive batch (B) normalization [16], ReLu activation (A) [17] and convolution layers (C), referred to as BAC from here on. This is a well-known fact in computer vision that best models for solving complex tasks contain lots of residual connections [18]. This is backed up by some recent works with more theoretical ground [19] and proved to be useful in our project as well. Hence, we utilize residual connections in building sub-modules for encoder and decoder parts of the network. The data flow between BAC layers is reshaped by a linear $1 \times 1$ convolution layer when the dimensions do not match. This $1 \times 1$ convolution layers are $L_1$ regulated to encourage sparsity and stop uninformative data flow among BAC layers. To encourage the network towards further generalization and prevent overfitting, spatial dropout layers are used. This is shown to be more effective than vanilla dropout in working with highly correlated spatial data such as images [20]. The overall architecture is shown in figure (2).

## 3.3. Adaptive thresholding inference

Deep convolutional neural networks can predict a heat map for each input. However, one of the main problems for treating these numbers as probabilities is that there are no confidence intervals or even weaker statistics such as standard deviation of these figures in order to provide an estimate about the confidence of the network belief. There have been some attempts to alleviate this problem by performing dropout at test time [2]. In this work, rather than injecting noise with unknown dynamics to the network internal representation, we propose a novel approach to introduce a metric for network belief. To this end, we sample from a linear manifold that the input image lies on, using random affine transformations. Having computed these samples and their corresponding heat maps, for each pixel, we compute median over all predictions for that pixel. Throughout the experiments, we found that median is more robust to outliers than the mean value. Alongside the median value, standard deviation for each pixel is computed as well. Now we define an adaptive threshold $D_i$, for each pixel depending on its statistics using equation (4):

$$D_i = B_i + \gamma \sigma_i \qquad (4)$$

where $B_i$ is the baseline score for classifying the pixel $i$ as foreground and $\sigma_i$ is its standard deviation. $\gamma$ is a hyper parameter of this stage which encodes how much weight we give to variance for increasing baseline $B_i$. Setting $\gamma$ to zero and fixing $B_i$ will reduce the algorithm to classical fixed hard thresholding. With $D_i$ in hand for each pixel, the final segmentation mask $S$, is computed using a simple pixel-wise comparison as shown by equation (5):

$$S_i = \begin{cases} 1 & if\ m_i > D_i \\ 0 & otherwise \end{cases} \qquad (5)$$

where $S_i$ is the final segmentation value and $m_i$ is the median of heat maps for pixel $i$. There are a lot of ways to improve this thresholding mechanism, inspired by ideas from classical image processing, such as Sauvola algorithm [13]. One suggestion is to adapt $\gamma$ based on the statistics of each pixel neighborhood.

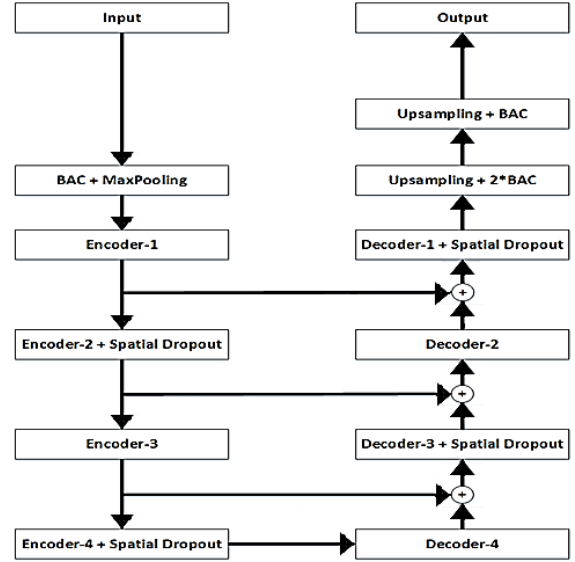

Figure 2. Overall deep fully convolutional network's architecture.

## 4. EXPERIMENTAL RESULTS

We use the York dataset of cardiac MRI sequence [1] for training and evaluation. Dataset consists of 33 patients, for each of which, there are 20 MRI series with 8 to 15 slices. Each slice is a single channel image with spatial size of

$256 \times 256$. In total, dataset contains 7980 MRI images, while only 5011 of them have segmentation ground-truth. In training and testing phases, we solely use the frames with segmentation labels. However, using other unsegmented frames might be beneficial for computing temporal structure or unsupervised pre-training.

Comparison to other recently proposed methods is presented in section 4.2, which shows superiority of our proposed method.

We also conducted another interesting experiment by using fully connected conditional random fields (CRF) [21], in order to refine the final segmentation map. The main purpose of this experiment was to compare the performance of our adaptive thresholding algorithm to a more sophisticated method such as the probabilistic graphical model CRF. We incorporated the median and standard deviation as computed in section 3.3 in the unary term of the CRF and tried to tune other hyper parameters using cross validation and random search. However, we did not observe any improvements over our adaptive thresholding method. This result suggests that simpler and more interpretable models cannot be overlooked. Indeed, these techniques can be complementary to more capable methods including deep neural networks.

### 4.1. Training and parameters tuning

As there are no official training and validation splits available for this dataset, fair comparison with other methods is hard. In order to find a way around this problem, we repeat our experiment 10 times with random splits and report the average results. During the training phase for each experiment, the dataset is shuffled first and then decomposed into two non-overlapping sets. The first set is used for evaluation and the second one is set aside for training the deep neural network. Frames from each patient MRI scan could only be present in one of the sets. The test set consists of frames from five randomly selected patients and other 28 patients are used in training phase. To save computational power, we resize the 256×256 frames to 128×128.

The deep FCN (c.f. section 3.2) is trained for 3000 epochs using Adam [22] with initial learning rate of $10^{-3}$. We want to get the best performance in terms of Dice coefficient, so we define a new loss function to directly incorporate Dice coefficient as follows:

$$Loss(y_t, y_p) = BCE(y_t, y_p) - e^{(1+Dice(y_t, y_p))} \qquad (6)$$

where $y_t$ is the ground truth and $y_p$ is the prediction of the model. BCE is the binary cross-entropy that is widely used in binary classification problems. The use of exponential function here produces larger gradients for optimization and hence the problem of vanishing gradients is reduced to some extent. Regularization hyper parameters such as spatial dropout rate and $L_1$ regularization for residual connection in encoder and decoder sub modules were set to 0.5 and $10^{-5}$

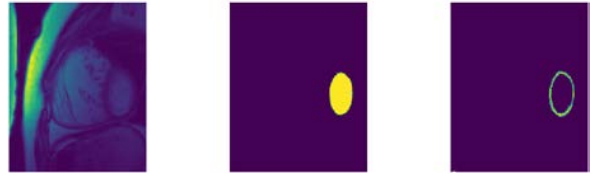

Figure 3. From left to right: input image, median prediction, and standard deviation in predictions for each pixel.

respectively. Data augmentation such as vertical/horizontal flips, rotations, translations and zooming are also applied during training in order to achieve higher generalization and reduce over fitting effects due to small size of our training set. 0.5

For hyper parameters of random affine transformation and adaptive thresholding in equations (1) to (4), we first set $T = 20\ px$ and $R = 20°$. Then, we optimize $\gamma$ using cross validation. Values in range $[10^{-2}, 0.5]$ have given provide good performance. We set $\gamma = 0.1$ and $B_i = 0.5$ in equation (4) for our experiments.

### 4.2. Results and comparisons

Figure (3) shows the results of using median and standard deviation in prediction. An important observation depicted in figure (3) is that the highest uncertainty shown by the model is about the boarders of left ventricle. This phenomenon has been reported by previous works [2]. The overall Dice score values [9] are shown in Table (1) As can be observed, the proposed FCN with adaptive thresholding results in the best Dice score.

Table 1. Models evaluation using Dice coefficient

| Method | Dice(%) |
|---|---|
| Fast-Segment [9] | 85.9 |
| Proposed (FCN Only) | 89.66 |
| Proposed (FCN + CRF) | 89.80 |
| Proposed (FCN+Adaptive Thresholding) | 90.2 |

### 5. CONCLUSION

In this paper we introduced a method to measure uncertainty in predictions of deep convolutional neural networks by sampling from a linear manifold the input image lies on. Sampling proceeds with generating random affine transformation matrix and applying it on input images. Having computed multiple outputs and the approximate uncertainty for each pixel, we used an adaptive thresholding algorithm to obtain the final segmentation map. The architecture and loss function of the fully convolutional network have also been engineered to improve training phase and faster convergence as well as model accuracy. Further investigation for modelling uncertainty of the deep CNN in their predictions is crucial to expand our understanding about internal mechanisms of these networks.